# Intra-and-Inter-Constraint-based Video Enhancement based on Piecewise Tone Mapping

Yuanzhe Chen, Weiyao Lin, Chongyang Zhang, Zhenzhong Chen, Ning Xu, and Jun Xie

***Abstract*— Video enhancement plays an important role in various video applications. In this paper, we propose a new intra-and-inter-constraint-based video enhancement approach aiming to 1) achieve high intra-frame quality of the entire picture where multiple region-of-interests (ROIs) can be adaptively and simultaneously enhanced, and 2) guarantee the inter-frame quality consistencies among video frames. We first analyze features from different ROIs and create a piecewise tone mapping curve for the entire frame such that the intra-frame quality of a frame can be enhanced. We further introduce new inter-frame constraints to improve the temporal quality consistency. Experimental results show that the proposed algorithm obviously outperforms the state-of-the-art algorithms.**

## I. Introduction

Video services are of increasing importance in many areas including communications, entertainment, healthcare, and surveillance [1, 4, 8-10]. However, in many applications, the quality of video service is still hindered by several technical limitations such as poor lightening conditions, bad exposure level, and unpleasant skin color tone [8-9]. Thus, it is crucial to enhance the perceptual quality of videos. In this paper, we focus on two issues for video enhancement:

### *A. Intra-frame quality enhancement with multiple region of interests (ROIs)*

Since a frame may often contain multiple regions of interests, it is desirable for the enhancement algorithm to achieve high intra-frame quality of the entire picture where multiple region-of-interests (ROIs) can be adaptively and simultaneously enhanced. However, many existing works [3-4,11-14] focus on enhancing the intra-frame quality of a picture based on some pre-defined global metrics. Since these methods do not consider region differences within an image, they cannot guarantee all the important regions inside the image to be enhanced properly.

Other approaches [7-8,12] identify and improve the perceptual quality of some specific regions in an image. Shi et al. [7] and Battiato et al. [12] designed exposure correction methods [3-4] to perform the exposure correction based on the features of some relevant regions. Liu et al. [8] proposed a learning-based color tone mapping method to conduct global color transfer by turning the color statistic of the face region according to a pre-trained set. Although these methods can improve the quality of some specific regions such as the face, the quality of other regions may be deteriorated. Thus, the overall perceptual quality may not be appealing.

There are some approaches which try to enhance multiple regions simultaneously. Some methods [11] adapt enhancement strategy according to the local neighborhood characters around each pixel. Although these methods suitably consider the image local characters, they are still not region-based methods since the region-of-interests can not be specified. Other methods [9-10] enhance multiple regions of interests within an image by segmenting the image into regions and performing color tone mapping for each region, respectively. However, these tone mapping methods are highly dependent on the segmentation results and may result in 'fake' edges between different regions.

### *B. Inter-frame quality enhancement among frames*

Most of the existing enhancement algorithms only focus on improving the intra-frame qualities within a single frame or an image. They are not suitable for enhancing videos since the inter-frame quality consistencies among frames are not considered. Some state-of-the-art algorithms can be extended for enhancing inter-frame qualities under some specific applications. For example, Liu et al. [8] propose a learning-based method for video conferencing where frames share the same tone mapping function if their backgrounds do not change by much. Although this method can achieve good inter-frame quality in video conferencing scenarios, it cannot be applied to other scenarios if the video backgrounds or contents change frequently. Toderici et al. [13] introduced a temporally-coherent method by combining the frame feature and the shot feature to enhance a frame. Their method can effectively enhance both the shot-change frames and the regular frames. However, this method relies on the performance of shot detector and fails to suitably enhance the inter-frame quality within a shot. Furthermore, Sun et al. [1] use additional hardware to compensate for the lighting condition for keeping consistency. Since this method has specific system requirements, it cannot be easily applied in other applications. Therefore, it is desirable to develop a more generalized algorithm which can handle the inter-frame quality enhancement of various videos.

In this paper, a new intra-and-inter-constraint-based (A+ECB) algorithm is proposed. The proposed algorithm analyzes features from different ROIs and creates a 'global' tone mapping curve for the entire frame such that different regions inside a frame can be suitably enhanced at the same time. Furthermore, new inter-frame constraints are introduced in the proposed algorithm to further improve the inter-frame qualities among frames. Experimental results demonstrate the effectiveness of our proposed algorithm.

The rest of the paper is organized as follows: Section II describes the motivation as well as the detailed process of our

Y. Chen, W. Lin, C. Zhang, N. Xu, and J. Xie are with the Department of Electronic Engineering, Shanghai Jiao Tong University, Shanghai 200240, China (e-mail: {yzchen0415, wylin, sunny_zhang, xn8812, jxie}@sjtu.edu.cn).

Z. Chen is with the MediaTek USA Inc., San Jose, CA 95134, USA (email: zzchen@ieee.org).

A+ECB algorithm. Section III shows the experimental results and Section IV concludes the paper.

## II. Intra-and-inter-constraint-based Video Enhancement

### A. Motivations

As mentioned, most existing approaches have various limitations in enhancing videos. Fig. 1 shows an example. Fig. 1 (c) and (e) are the enhanced results by the modified global histogram equalization algorithm [4] and a region-based method [8], respectively.

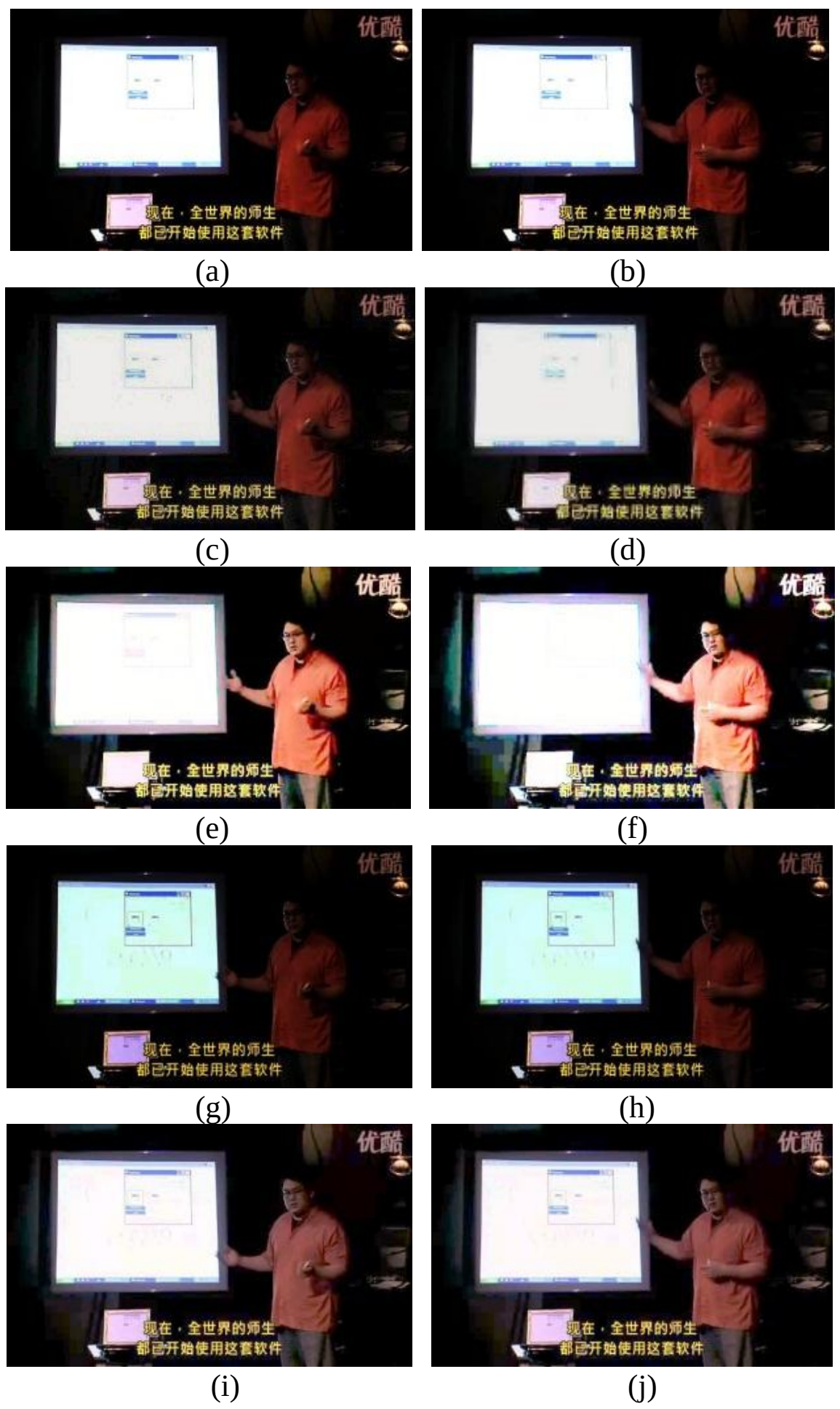

Fig. 1 (a)-(b) The original image, (c)-(d) Image enhanced by HEM [4], (e)-(f) Image enhanced by [8], (g)-(h) Enhanced image focusing on the screen region by [8], (i)-(j) The enhanced image by our A+ECB method. (Best view in color)

From Fig. 1 (c), we can see that since the image is enhanced based on a global contrast metric without considering the region difference, some of the important regions such as the face are not properly enhanced. Compared with Fig. 1 (c), since the region-based method [8] identifies the face region and performs enhancement accordingly, the visual quality of Fig. 1 (e) is much improved. However, since the tone mapping function trained from the face region will be applied to the entire image, the quality of some other regions such as the screen becomes poorer (e.g. the character on the screen in Fig. 1 (e) becomes difficult to tell). Similarly, when extending the region-based method [8] to enhance the screen region (as in Fig. 1 (g)), the quality of the face region becomes less appealing.

From the above discussions, we can have the following observations: (a) In order to achieve suitable enhancement results, features from regions of interests need to be considered. (b) It is desirable to enhance the entire frame 'globally' but with the consideration of different ROIs at the same time.

Therefore, we propose a new intra-frame-constraint-based (ACB) step for enhancing the intra-frame quality of a frame, as will be described in detail Section II-B.

Furthermore, as mentioned, most existing works [1-10] cannot effectively handle the inter-frame consistencies in a video. Although these methods may achieve proper visual qualities in each frame, the qualities among different frames (i.e., inter-frame qualities) may vary. This inconsistency may become severe when the algorithm adaptively adopts different enhancement parameters for different frames or when the color histograms of the original videos are changing quickly. Therefore, new algorithms which can handle inter-frame consistency are needed.

For the ease of description, we will discuss the idea of our algorithm based on the Histogram Equalization Modification-based (HEM) method [4]. However, it should be noted that the idea of our algorithm is general and it can be extended to other enhancement algorithms [1,7-10]. Furthermore, in order to make the description clear, in the rest of the paper, we will use boldface to represent vectors (i.e., color histograms) and use unboldface to represent scalars and functions.

The HEM method can be described as in Eqn. (1). Instead of using the histogram distribution to construct the tone mapping function directly, the method formulated a weighted sum of two objectives as:

$$\boldsymbol{h} = \arg\min_{\boldsymbol{h}^*} \left( \left| \boldsymbol{h}^* - \boldsymbol{e} \right|^2 + \lambda \cdot \left| \boldsymbol{h}^* - \boldsymbol{u} \right|^2 \right) \quad (1)$$

where $\boldsymbol{u}$ is the uniform distribution, $\boldsymbol{h}$ is the desired color histogram, $\boldsymbol{h}^*$ the possible candidate of $\boldsymbol{h}$, and $\boldsymbol{e}$ is the color histogram by traditional HE. $\lambda$ is a parameter balancing the importance between $\boldsymbol{u}$ and $\boldsymbol{e}$ [4]. From eqn. (1), the desired $\boldsymbol{h}$ can be achieved by:

$$\boldsymbol{h} = \left( \frac{1}{1+\lambda} \right) \cdot \boldsymbol{e} + \left( \frac{\lambda}{1+\lambda} \right) \cdot \boldsymbol{u} \quad (2)$$

Based on Eqn. (2), a tone mapping function can be calculated which enhances the original image to the desired color histogram $\boldsymbol{h}$ [4].

The basic idea of the HEM method is that by introducing another constraint (i.e., $|\boldsymbol{h}^* - \boldsymbol{u}|^2$), the unnatural effects in the HE histogram can be effectively reduced. However, the HEM method is still an intra-frame-based method which does not consider the temporal continuities among frames.

In order to handle inter-frame consistency, we can extend Eqn. (1) by including an additional inter-frame constraint by:

$$\boldsymbol{h} = \arg\min_{\boldsymbol{h}^*} \left( \left| \boldsymbol{h}^* - \boldsymbol{e} \right|^2 + \lambda \cdot \left| \boldsymbol{h}^* - \boldsymbol{u} \right|^2 + \gamma \cdot \left| \boldsymbol{h}^* - \boldsymbol{h}_{t-1} \right|^2 \right) \quad (3)$$

where $\boldsymbol{h}_{t-1}$ is the desired color histogram of the previous frame $t-1$ and $\gamma$ is another balancing parameter handling the importance of the inter-frame constraint. These parameter values would directly relate to the final enhancement results. In our experiment, $\lambda$ and $\gamma$ are set to be 2 and 3 respectively based on the experimental statistics. Note that these parameters and the other parameters in Eqns (4)-(8) can also be selected in an automatic way by selecting a set of parameters that maximizes the objective quality measurements in a suitable image database [11].

Therefore, we propose a new inter-frame-constraint-based (ECB) step for handling inter-frame constraints. The proposed ECB step will be described in detail in the following.

### *B. The Intra-and-Inter-Constraint-Combined Algorithm*

The framework of our intra-and-inter-constraint-based (A+ECB) algorithm can be described as in Fig. 2. In Fig. 2, an input frame is first enhanced by the proposed ACB step for improving the intra-frame quality. Then, the resulting frame will be further enhanced by the proposed ECB step for handling the inter-frame constraints. The ACB step and the ECB step will be described in detail in the following.

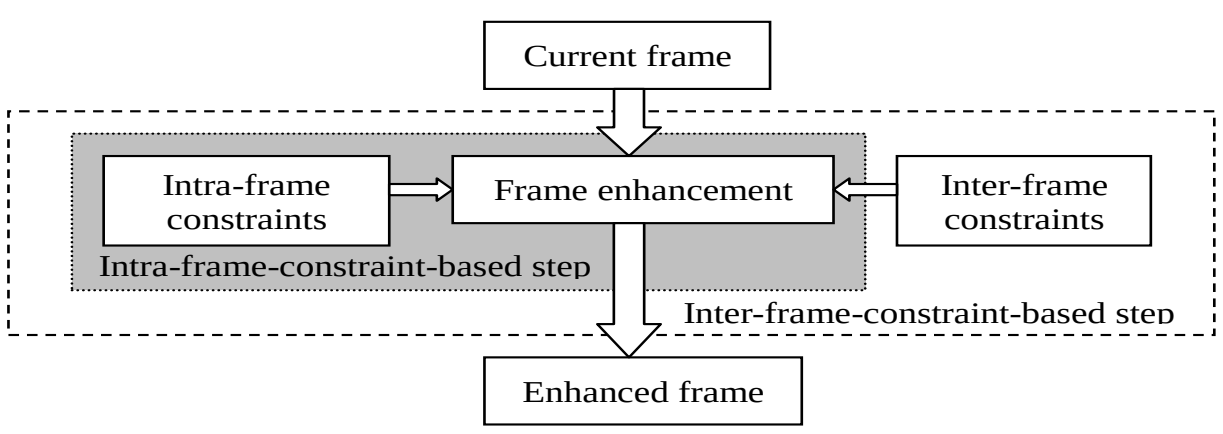

Fig. 2 The framework of the proposed A+ECB algorithm.

#### *1) The intra-frame-constraint-based step*

The process of the ACB step can be further described by Fig. 3. In Fig. 3, multiple ROIs are first identified from the input video frame. In this paper, we use video conferencing or video surveillance as example application scenarios and identify regions of interests (such as human faces, screens, cars, and whiteboards) based on an Adaboost-based object detection method [6]. Other object detection and saliency detection algorithms can also be adopted to obtain the ROIs. Note that since our algorithm creates a global tone-mapping curve for enhancement, these regions do not need to be perfectly extracted, as will be shown in the experimental results.

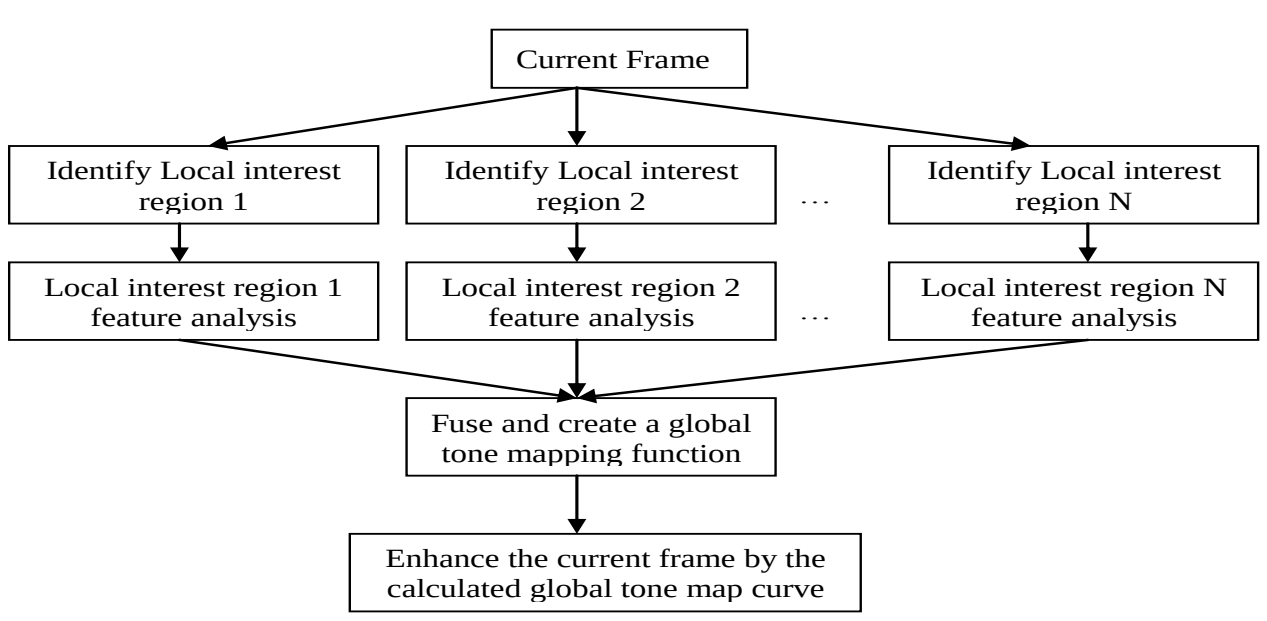

Fig. 3 The process of the ACB step.

After extracting and analyzing the features from these regions of interests, a global tone mapping curve is created by fusing these features from different regions. Finally, the enhanced frame by this global tone mapping curve can simultaneously provide appealing qualities for different regions of interests. It should be noted that the creation of the global tone-mapping curve is the key part of our algorithm. Therefore, in the rest of this section, we will focus on describing this part.

In order to create global tone-mapping curves, features need to be first extracted and analyzed for each region of interest. In this paper, we utilize a simple but effective method by extracting the mean $m_{Ri,j}$ and the standard deviation $\sigma_{Ri,j}$ for each ROI $R_i$ and for each color channel $j$ (we use R-G-B color channels and perform enhancement in each channel independently). However, note that our algorithm is general and the feature extraction as well as the color channel processing modules can also be implemented by more sophisticated ways. For example, if one ROI include multiple major colors, we can also view each major color region as a 'sub-ROI' and pre-fuse these sub-ROI features before fusing with other ROIs. Furthermore, the correlation constraints among color channels can also be included when performing enhancement for each color channel [4, 5, 9]. For the ease of description, we focus on discussing the global tone-mapping-curve creation from only two ROIs (i.e., $R_A$ and $R_B$) in this paper. The global curve creation from more ROIs can be easily extended in an iterative way (i.e., fuse two ROIs at each time and then view the fused ROIs as an entire ROI for later fusion). The ACB strategy for fusing multiple ROI features and for creating the global tone mapping curve can be described by Eqn. (4).

$$f_{g,j}(x)=\begin{cases} f_{g,j}^{PB}(x) & if\ \left(m_{RA,j}+\rho\cdot\sigma_{RA,j}\right)-\left(m_{RB,j}-\rho\cdot\sigma_{RB,j}\right)<TH \\ f_{g,j}^{FB}(x) & if\ \left(m_{RA,j}+\rho\cdot\sigma_{RA,j}\right)-\left(m_{RB,j}-\rho\cdot\sigma_{RB,j}\right)\ge TH \end{cases} \qquad (4)$$

where $f_{g,j}(x)$ is the fused global tone mapping curve for intra-frame enhancement in color channel $j$. $x$ is the color level value. $f_{g,j}^{PB}(x)$ is the tone mapping curve by the piecewise-based strategy for color channel $j$ and $f_{g,j}^{FB}(x)$ is the curve by the factor-based strategy. $m_{Ri,j}$ and $\sigma_{Ri,j}$ are the mean and the standard deviation for ROI $R_i$ in color channel $j$. In this paper, we assume that $m_{RA,j} < m_{RB,j}$. $TH$ is a threshold and $\rho$ is a parameter for reflecting the separability of the two ROIs. In our experiment, TH is set to be 50 and $\rho$ is set to be 1 based on the experimental statistics.

From Eqn. (4), we can see that if the features of the two ROIs $R_A$ and $R_B$ are different from each other (i.e., the pixel values for regions $R_A$ and $R_B$ seldom overlap and can be roughly separated: $(m_{RA,j}+\rho\cdot\sigma_{RA,j})-((m_{RB,j}-\rho\cdot\sigma_{RB,j})<TH)$), the fused global curve $f_{g,j}(x)$ will take the piecewise-based form $f_{g,j}^{PB}(x)$. Otherwise, $f_{g,j}(x)$ will take the factor-based form: $f_{g,j}^{FB}(x)$.

The basic idea of our piecewise-based curve-fusion strategy $f_{g,j}^{PB}(x)$ can be described as follows: if the features of the two ROIs $R_A$ and $R_B$ are different from each other (i.e., the pixel values for $R_A$ and $R_B$ seldom overlap and can be roughly separated), then the fused global tone mapping curve $f_{g,j}(x)$ can be divided into two parts where each part can be tuned to suitably enhance the quality of its corresponding ROI. More specifically, $f_{g,j}^{PB}(x)$ can be described by:

$$f_{g,j}^{PB}(x)=\begin{cases} f_{RA,j}(x) & if\ x\in[0,\ P_c] \\ f_{RB,j}(x) & if\ x\in(P_c,\ 255] \end{cases} \qquad (5)$$

where $f_{RA,j}(x)$ and $f_{RB,j}(x)$ are the piecewise tone mapping parts corresponding to the local ROIs $R_A$ and $R_B$, respectively. $P_c$ is the conjunctive point for the two parts. In this paper, $f_{RA,j}(x)$ and $f_{RB,j}(x)$ are modeled as piecewise cubic spline functions. The piecewise part $f_{RA,j}(x)$ can be derived from the constraints in Eqns (6)-(7). The second piecewise part $f_{RB,j}(x)$ can be calculated in a similar way.

$$\begin{cases} f_{RA,j}(0)=0 \\ f_{RA,j}(m_{RAj}-\alpha\cdot\sigma_{RAj})=k_1\cdot f_{A,j}(m_{RAj}-\alpha\cdot\sigma_{RAj})+(1-k_1)\cdot f_{B,j}(m_{RAj}-\alpha\cdot\sigma_{RAj}) \\ f_{RA,j}(P_c)=k_2\cdot f_{A,j}(P_c)+(1-k_2)\cdot f_{B,j}(P_c) \end{cases} \quad (6)$$

where $f_{A,j}(x)$ is the local tone mapping curve for enhancing the ROI $R_A$ and it can be easily calculated by some learning-based methods according to the extracted features in each ROI (i.e., $m_{RA,j}$ and $\sigma_{RA,j}$) [8]. $k_1$ ($0 \le k_1 \le 1$) is the parameter to determine the level of the similarity between $f_{RA,j}(x)$ and $f_{A,j}(x)$. $f_{RA,j}(x)$ will be closer to $f_{A,j}(x)$ around ROI $R_A$'s major color region (i.e., around $m_{RA,j}$) for larger $k_1$ values. $k_2$ ($0 \le k_2 \le 1$) is a parameter to determine the pixel values around the intersection region between the two piecewise parts $f_{RA,j}(x)$ and $f_{RB,j}(x)$. Similar to $k_1$, $f_{RA,j}(x)$ will be close to $f_{A,j}(x)$ around the intersection region for large $k_2$ values, and will be close to $f_{B,j}(x)$ for small $k_2$ values. In our experiment, $k_1$ is set to be 0.9 since we would like $f_{RA,j}(x)$ to be close to $f_{A,j}(x)$ around the $R_A$'s major color region. $k_2$ is set to be *0.5* such that the color information of $R_A$ and $R_B$ can be equally enhanced at $f_{RA,j}(P_c)$.

Furthermore, in order to smooth the shape of the fused global curve $f_{g,j}^{PB}(x)$ and to prevent quick saturation at two ends, we also add the following derivative constraints on $f_{RA,j}(x)$:

$$\begin{cases} f'_{RA,j}(0)=0.5\cdot\dfrac{f_{RA,j}(m_{RAj}-\alpha\cdot\sigma_{RAj})}{m_{RAj}-\alpha\cdot\sigma_{RAj}} \\ f'_{RA,j}(m_{RAj}-\alpha\cdot\sigma_{RAj})=k_1\cdot f'_{A,j}(m_{RAj}-\alpha\cdot\sigma_{RAj})+(1-k_1)\cdot f'_{B,j}(m_{RAj}-\alpha\cdot\sigma_{RAj}) \\ f'_{RA,j}(P_c)=k_3\cdot\dfrac{f_{RA,j}(P_c)-f_{RA,j}(m_{RAj}-\alpha\cdot\sigma_{RAj})}{P_c-(m_{RAj}-\alpha\cdot\sigma_{RAj})} \end{cases} \quad (7)$$

where $k_3$ is a parameter to determine the shape of $f_{RA,j}(x)$ in higher intensities. By introducing the constraints in Eqn. (7), the smoothness and the non-decreasing properties for $f_{RA,j}(x)$ can be guaranteed around the starting region (i.e., 0), the major color region of ROI $R_A$ (i.e., $m_{RA,j}-\alpha\cdot\sigma_{RA,j}$ ), and the intersection region between $f_{RA,j}(x)$ and $f_{RB,j}(x)$ (i.e., $P_c$). The value of $k_3$ is determined by the values of $f_{A,j}(P_c)$ and $f_{B,j}(P_c)$. If $f_{A,j}(P_c) > f_{B,j}(P_c)$, $k_3$ is a real number between 0 and 1. Otherwise, $k_3$ is a real number between 1 and $+\infty$. In our experiment, $k_3$ is set to be 0.5 or 1.5 in these two cases respectively to make the gradient at $f_{RA,j}(P_c)$ smooth with both the $f_{RB,j}(x)$ curve part and the early part of $f_{RA,j}(x)$ around $m_{RA,j}$.

The conjunctive point $P_c$ in Eqn. (5)-(7) can be obtained by:

$$P_c=\begin{cases} \dfrac{(m_{RA,j}+\alpha\cdot\sigma_{RA,j})+[m_{RB,j}-\alpha\cdot\sigma_{RB,j}+(n_{A,j}-n_{B,j})\cdot\sigma_{RA,j}]}{2} & n_{A,j}\ge n_{B,j} \\ \dfrac{[m_{RA,j}+\alpha\cdot\sigma_{RA,j}-(n_{B,j}-n_{A,j})\cdot\sigma_{RB,j}]+(m_{RB,j}-\alpha\cdot\sigma_{RB,j})}{2} & n_{A,j}< n_{B,j} \end{cases} \quad (8)$$

where $n_{i,j}$ ($i$=$A$ or $B$) denotes the feature difference between ROI $R_i$ and the other ROI for channel *j*. They are calculated by the ratio between the 'pixel value overlapping area of the two regions' and 'the variance $\sigma_{Ri,j}$ of ROI $R_i$'. In this paper, the pixel value overlap- ping area is calculated based on the 3σ areas [2] of the two region of interests (i.e., $\alpha=3$ in Eqns (6)-(8)). Note that $n_{i,j}$ can also be viewed as the measure of the pixel-value separability between the ROIs and it will equal to *0* if there is no overlapping area and the pixel-value of the two ROIs can be completely separated.

Fig. 4 (a) shows an example of a fused global tone mapping curve by the piecewise-based strategy. From Fig. 4 (a), we can see that with the piecewise-based strategy, the final tone mapping curve (i.e., the black solid curve) is composed of two parts where the first part is close to the local curve of $R_A$ around its $3\sigma$ area and the second part is close to the local curve of $R_B$ around $R_B$'s $3\sigma$ area. By this way, both ROIs can be properly enhanced by the created global tone mapping function.

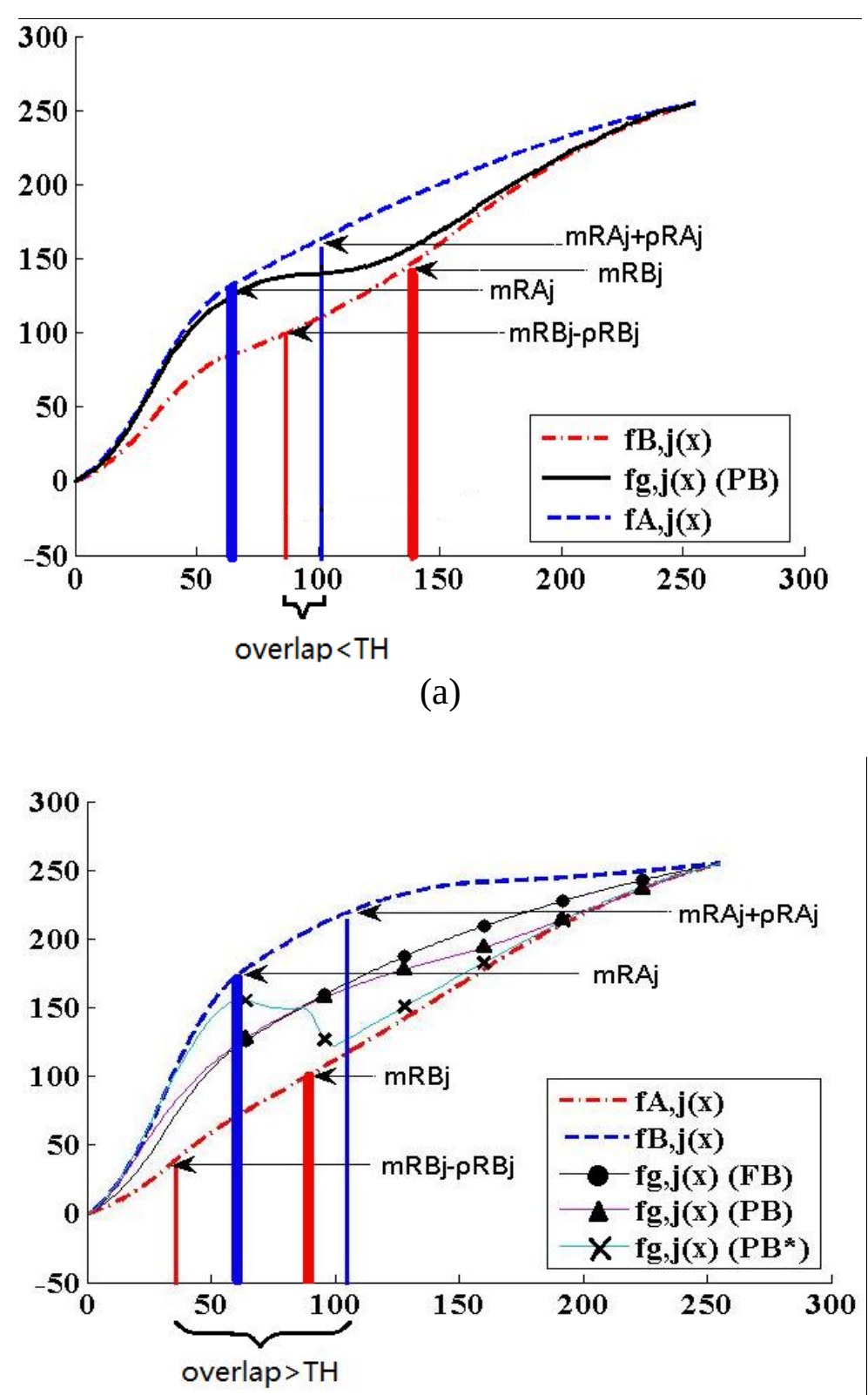

Fig. 4 (a) The fusion for different-ROI-feature case for the red color channel, including the local tone mapping curves for ROI $R_A$ ($f_{A,j(x)}$, blue dotted curve), ROI $R_B$ ($f_{B,j(x)}$, red dash-dot curve), and the fused global mapping curve by the piecewise strategy ($f_{g,j(x)}$, black solid curve). (b) The fusion for similar-ROI-feature case for the red color channel, including the local tone mapping curves for ROI $R_A$ (blue dotted curve), ROI $R_B$ (red dash-dot curve), the fused global mapping curve by the piecewise strategy without constraint (the blue cross-mark curve), the fused global mapping curve by the piecewise strategy with constraint (the pink triangle-mark curve), and the fused global mapping curve by the factor-based strategy(the black diamond curve).

Although the proposed piecewise-based strategy is effective in handling the difficult problem of simultaneously enhancing multiple ROIs which have large differences in pixel value statistics (i.e., features), it is not suitable in cases when the ROIs have similar features, as it may create unsmooth tone mapping curves. In this case, $f_{g,j}(x)$ will take the factor-based form $f_{g,j}^{FB}(x)$, by:

$$
P_c = \begin{cases} \dfrac{(m_{RA,j} + \alpha \cdot \sigma_{RA,j}) + [m_{RB,j} - \alpha \cdot \sigma_{RB,j} + (n_{A,j} - n_{B,j}) \cdot \sigma_{RA,j}]}{2} \\ \dfrac{[m_{RA,j} + \alpha \cdot \sigma_{RA,j} - (n_{B,j} - n_{A,j}) \cdot \sigma_{RB,j}] + (m_{RB,j} - \alpha \cdot \sigma_{RB,j})}{2} \end{cases} \tag{9}
$$

where $x$ is the color level value, $\lambda_j$ ($0 \le \lambda_j \le 1$) can be obtained by solving a optimization problem as:

$$
\lambda_j = \arg\min_{\lambda_j^*} \{ \| f_{g,j}^{FB}(x) - f_{A,j}(x) \|_{x \in [m_{RA,j} - \alpha \cdot \sigma_{RA,j}, m_{RA,j} + \alpha \cdot \sigma_{RA,j}]} + \| f_{g,j}^{FB}(x) - f_{B,j}(x) \|_{x \in [m_{RB,j} - \alpha \cdot \sigma_{RB,j}, m_{RB,j} + \alpha \cdot \sigma_{RB,j}]} \} \tag{10}
$$

From Eqn (9), we can see that the factor-based strategy creates the global tone mapping function by embedding the fusion information into the weighting factor $\lambda_j$. By this way, both of the ROIs $R_A$ and $R_B$ can be suitably enhanced.

Fig. 4 (b) and Fig. 5 are an example to show the necessity of using the factor-based strategy when ROI features are similar. In Fig. 5, (a) is the original image where we identify two ROIs for enhancement: the left car and the road; (b) is the enhancement results by the piecewise-based strategy without derivative constraints (i.e., creating the piecewise curve only based on Eqn. (7) and without Eqn. (8)); (c) is the enhancement results by the regular piecewise-based strategy (i.e., creating the curve based on both Eqn. (7) and (8)); And (d) is the result by the factor-based strategy. Fig. 4 (b) shows the corresponding tone mapping curves for Fig. 5 (b)-(d), respectively. Furthermore, the local tone mapping curves of the two ROIs in Fig. 5 (a) are also plotted in Fig. 4 (b) (the two dashed curves).

From Fig. 4 (b) and Fig. 5, we can see that since the color features of the two ROIs are similar (i.e., their mean values are very close to each other), the pixel values for the two ROIs have large overlapping areas. If we directly use the piecewise strategy without the derivative constraints in Eqn. (7), the created curve (the blue cross-mark curve) becomes unsmooth or even not monotonically increasing. Thus the enhancement result (Fig. 5 (b)) includes unnatural colors. If we include Eqn. (7)'s derivative constraints during piecewise strategy, the created curve (the pink triangle-mark curve) becomes smoother and more reasonable. Its corresponding enhancement result (Fig. 5 (c)) is much improved from (b). However, since the features of the two ROIs are close to each other, the tone mapping curve by the piecewise strategy is still less effective in creating a satisfactory curve balancing both ROIs. In Fig. 5 (c), the street regions are tuned to be green since the piecewise curve around these color values is unsuitably tuned to favor the car while neglecting the effect of the street. Compared to the piecewise curve, the tone mapping curve created by the factor-based method can suitably balance both the ROIs and creates the most appealing result (Fig. 5 (d)).

It should be noted that although the piecewise strategy is less effective for similar ROI feature cases, it will create better results than the factor-based strategy when ROI features are different. This is because when ROI features are different, the colors from different ROIs seldom overlap, and thus the tone mapping for the color values in one ROI will become less effective to the other. In this case, the piecewise strategy has more flexibility to make full use of the color resource to enhance multiple ROIs satisfactorily. This will be further demonstrated in the experimental results.

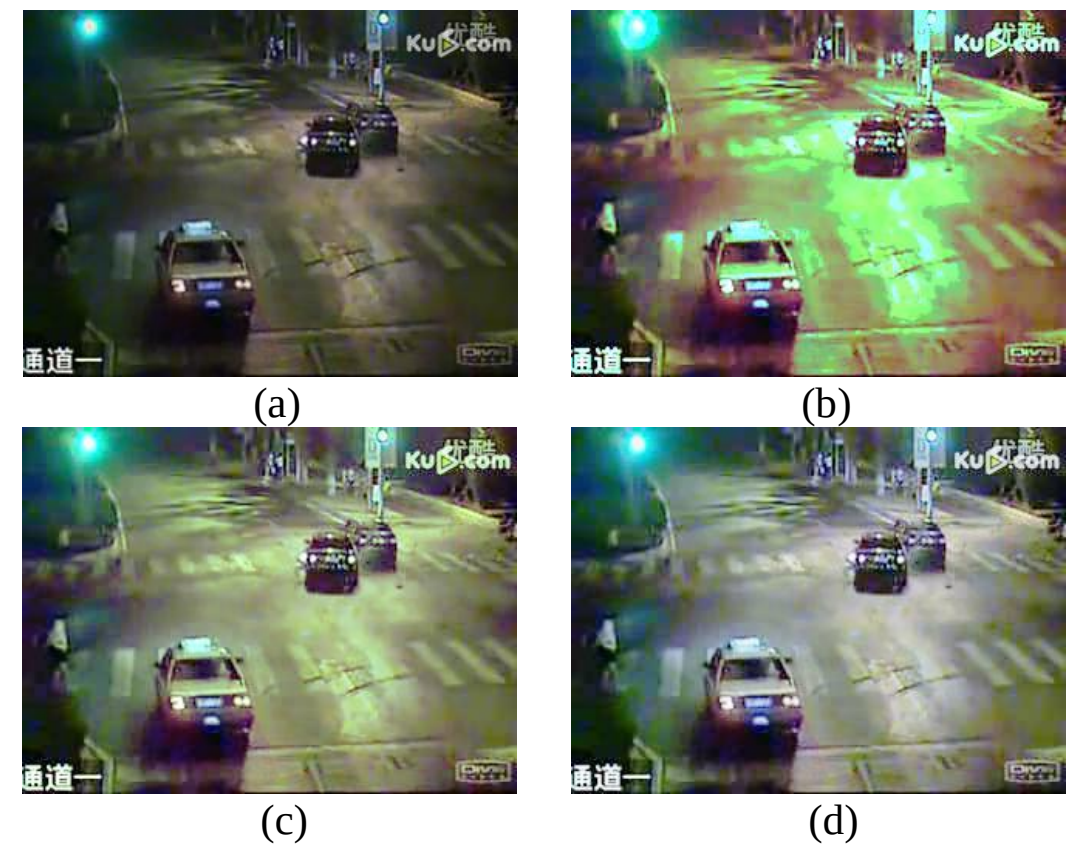

Fig. 5 (a) The original image, (b) The image enhanced by the piecewise strategy without constraint factor-based strategy, (c) The image enhanced by the piecewise-based strategy with constraint, and (d) The image enhanced by the factor-based strategy (best view in color).

*2) The inter-frame-constraint-based step*

The ECB step can be implemented by the HEM-based framework as mentioned in Eqn. (3). In our paper, besides Eqn. (3), we also propose another ECB step described by Eqn. (11). It should be noted that both Eqn. (3) and Eqn. (11) are based on the same inter-frame enhancement idea discussed in Section II.

$$
f_{g,j}^{A+E}(x) = (1 - \lambda_j^{EA}) \cdot f_{g,j}(x) + \lambda_j^{EA} \cdot f_{preg,j}^{A+E}(x) \tag{11}
$$

where $f_{g,j}(x)$ is the intra-frame global tone mapping curve from Eqn. (4). $f_{preg,j}^{A+E}(x)$ is the tone mapping curve by the A+ECB method in the previous frame. $\lambda^{EA}$ is the balancing parameter with the inter-frame constraints embedded, calculated by:

$$
\lambda_{EA} = \max\left( \arg\min_{\lambda_{EA}^*} \left| E(t) - E(t-1) \right|, \mathrm{LB} \right) \tag{12}
$$

where $E(t)$ is the entropy of frame $t$ and it can be calculated by:

$$
E = \sum_k -p(k) \cdot \log p(k) \tag{13}
$$

where $p(k)$ is the histogram value at bin $k$. Note that a lower-bound $LB$ is defined in Eqn. (12) to ensure that the inter-frame constraint can be effective in controlling the inter-frame consistencies. $LB$ is set to be 0.5 in our experiment. From Eqns (11)-(13), we can see that this ECB strategy embeds the inter-frame constraints in the balancing parameter $\lambda_{EA}$ such that results of the ACB step can be shifted to a relatively stable intensity level. Thus, the visual qualities in each frame can be kept coherent and the inter-frame discontinuity can be effectively reduced.

## III. Experimental results

### *A. Results for the intra-frame-constraint-based step*

Fig. 6 compares the enhancement results for different intra-frame enhancement methods. From Fig. 6, we can see that since the color of the two people are far different to each other, the learning-based method cannot properly enhance both faces simultaneously. As in Fig. 6 (b)-(c), when it enhances the face of one person, the quality of another person's face becomes unsatisfactory. Although by using the factor-based strategy in (d), the trade-off between the two faces can be improved, it is still less effective in creating a tone mapping curve for enhancing both ROIs. We can see that the face of the right person is still dark in (d). Comparatively, our ACB algorithm will select the piecewise strategy which calculates a fused piecewise global tone mapping function based on both regions. By this way, we can achieve satisfactory qualities in both faces, as shown in (e). Moreover, although the original video from each party may have large difference in illumination conditions, the enhancement results of different users are more coherent by our algorithm. This leads to better visual experience to users.

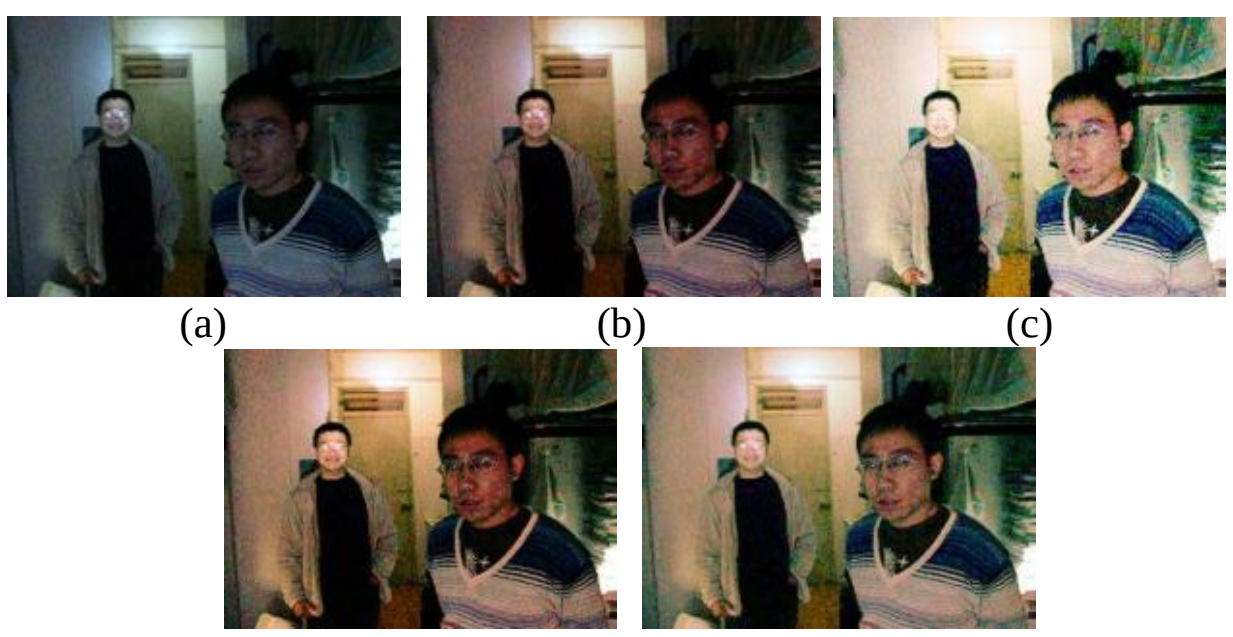

Fig . 6 (a) The original image, (b) and (c) The image enhanced by [8] based on the left and the right face region respectively, (d) The image enhanced by the factor-based strategy, (e) The image enhanced by our proposed ACB step with the piecewise strategy (best view in color).

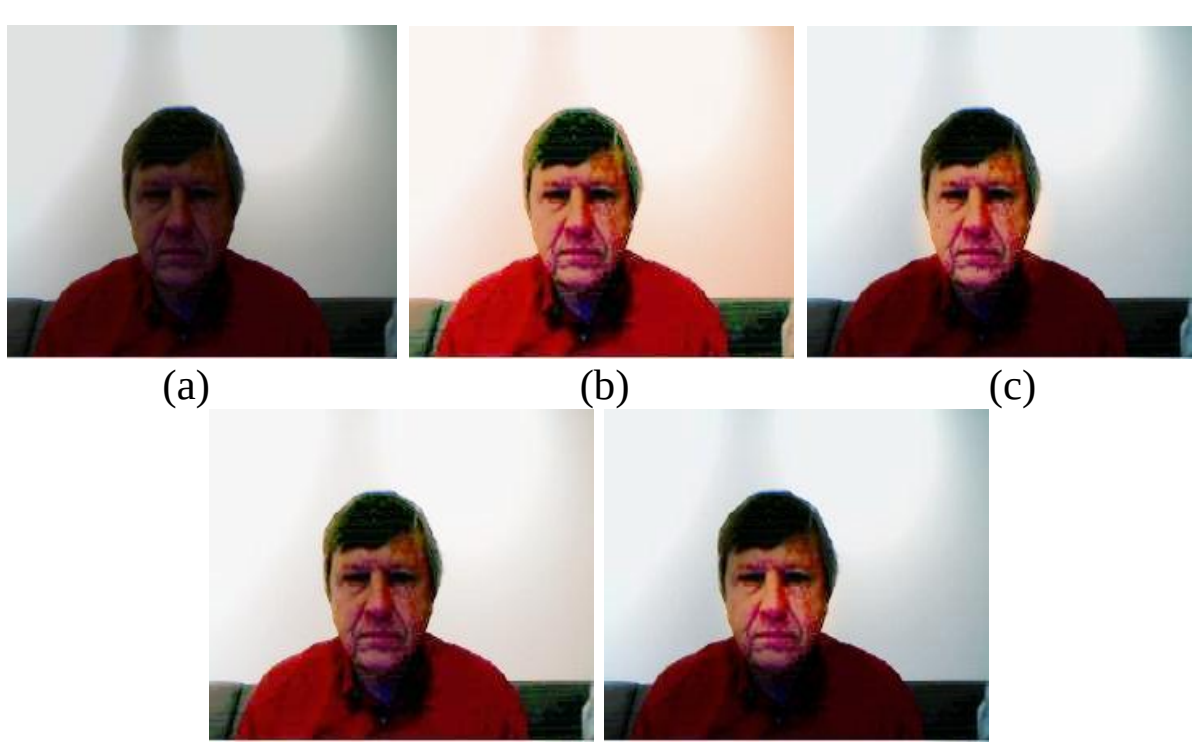

Fig. 7 (a) The original frame, (b) The frame enhanced by [8], (c) The image enhanced by directly putting different locally-enhanced regions together, (d) The image enhanced by the factor-based strategy, (e) The image enhanced by the proposed ACB step with the piecewise strategy (best view in color).

Furthermore, Fig. 7 shows another example. From Fig. 7, we can see that since the learning-based method performs enhancement only based on the color statistics in the face region, the color of the white background is not properly enhanced in (b) (i.e., it becomes red). Furthermore, although (c) properly enhances the background and the head, it creates fake edges around the person's head due to its local enhancement process (i.e., the red part around the head). Similar to Fig. 6, since the features of ROIs are different, the factor-based strategy in (d) still creates less appealing result as the background still turns a little bit red. Comparatively, the result in (e) is more appealing in both the face and the background regions while avoiding the fake edge effect at the same time.

### *B. Results for the inter-frame-constraint-based step*

In this section, we show results for our ECB step. Note that our ECB step is most effective in the following two application scenarios: (a) The original video is flicking or temporally inconsistent and the proposed algorithm can effectively reduce these inter-frame inconsistencies. (b) The original video sequence is temporally consistent. But after applying intra-frame enhancement methods on each frame, the inter-frame consistency decreases.

The example results for case (a) and (b) are shown in Figs. 8,10 and Fig. 1, respectively. The videos for Fig. 8 and 10 have inconsistent inter-frame qualities since they are captured under a lightning weather or a frequent light changing scenario. The ECB step in Fig. 8 and 10 is implemented by Eqn. (3).

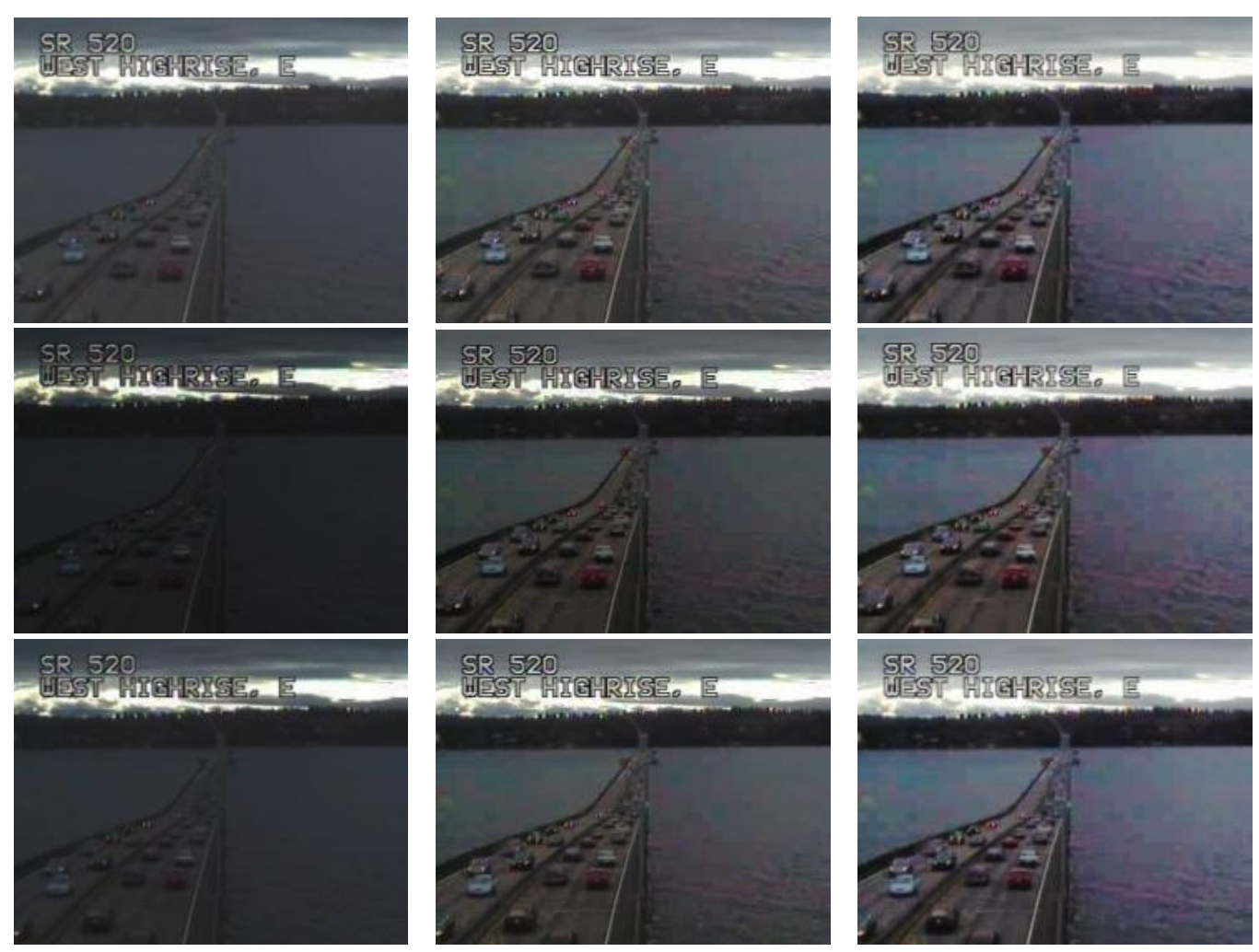

Fig. 8 Left column: original video sequence (correspond to the 3rd, 4th and 7th frames of the original video), middle column: enhanced video sequence by HEM [4], right column: enhanced video sequence by applying our inter-frame constraints (ECB).

Comparing the sequences in Fig. 8 and 10, the effect of our ECB algorithm is apparent. In the left columns of Fig. 8 and 10, the illumination flickers obviously in different frames due to the unstable light condition. Although the HEM method can properly improve the intra-frame quality in each frame (the middle columns in Fig. 8 and 10), the temporal inconsistency among frames still exists. Compared with the HEM method, the proposed ECB step can effectively improve both the intra-frame and the inter-frame qualities in the video. We can see from the right columns of Fig. 8 and 10 that the inter-frame inconsistency is properly eliminated and the contrast in each frame is also properly tuned to become visually appealing.

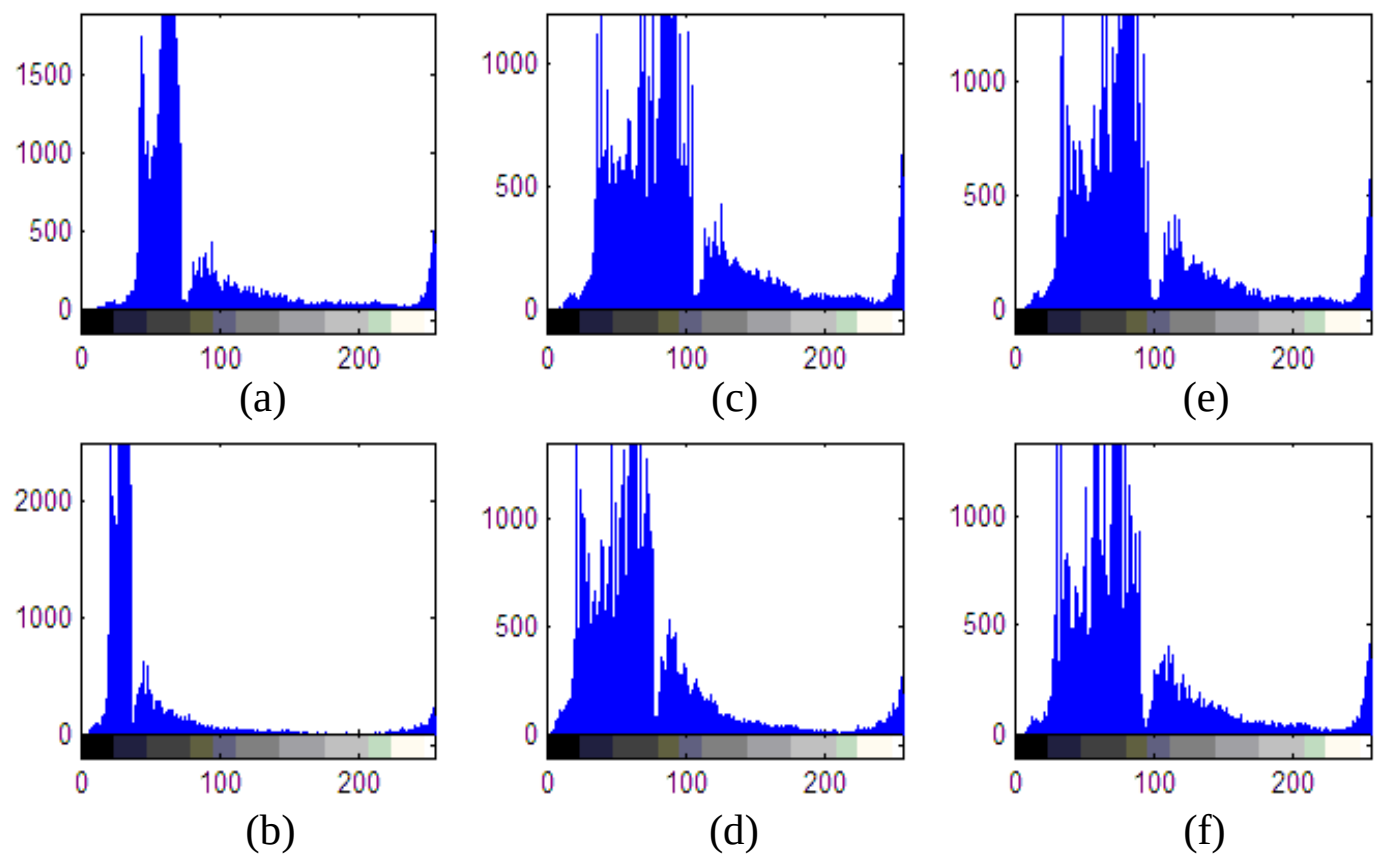

Fig. 9 Histograms of the first and second row images in Fig. 8.

Fig. 9 shows the corresponding histograms of the first and second row images in Fig. 8. From Fig. 9, we can see that due to the lightening weather condition, there is a clear distribution shift in the histograms of the original sequence (i.e., (a) and (b)). When enhanced by the HEM algorithm, the distribution shift is still obvious in (c) and (d). However, when enhanced by our ECB algorithm, the histogram distributions of the two continuous frames are tuned to have similar shapes and distributions. This also demonstrates the ability of our method in handling the inter-frame consistency.

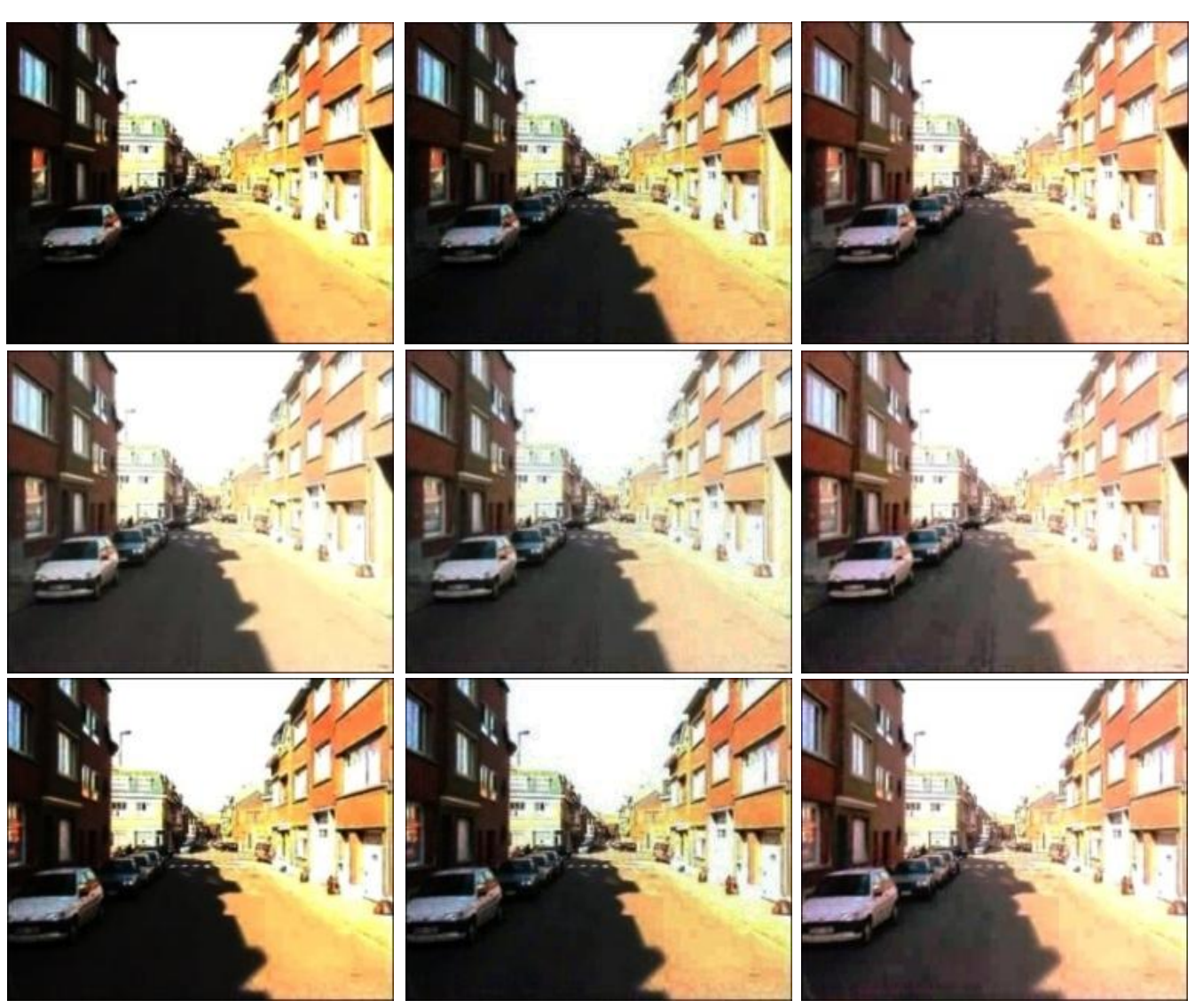

Fig. 10 Left column: Original video sequence, middle column: enhanced video sequence by HEM [4], right column: enhanced video sequence by applying our inter-frame constraints (ECB).

### *C. Results for combining intra-frame and inter-frame steps*

An example result of the proposed A+ECB algorithm is shown in Fig. 1. Similarly, when performing intra-frame enhancement only based on the face region (i.e., Fig. 1 (e)-(f)) or only based on the screen region ((g)-(h)), the other regions cannot be properly enhanced. However, by using our A+ECB algorithm both regions of interests can be enhanced properly. Furthermore, since the face region is not perfectly identified, we can see that the enhanced results in Fig. 1 (e) and (f) are not temporally consistent. However, by including our ECB step in (i)-(j), the inter-frame quality of the video can be guaranteed.

Table 1 compares the objective measurements for different methods. Table 1 compares the discrete entropy (*H*), temporal absolute mean brightness error ($TAMBE(\mu)$) [4], the standard deviance of the difference image between the neighboring frames ($TAMBE(\sigma)$) [4], and histogram- intersection-based temporal error (*HIBTE*) [14]. Normally, large *H* values reflect good intra-frame qualities while small $TAMBE(\mu)$, $TAMBE(\sigma)$, and *HIBTE* values imply good inter-frame qualities [4,14].

From Table 1, we can see that the intra-frame quality (i.e., H) of the ECB method is a little degraded from that of HEM. This is because by including the inter-frame constraints in ECB, the enhancement extent for each frame is limited in order to keep coherent with neighboring ones. However, this intra-frame quality degradation by ECB is small. Comparatively, the inter-frame quality improvements by ECB are obvious, which leads to an overall improvement of ECB from HEM. Similarly, although our ACB method can obviously improve the intra-frame quality of the video, its inter-frame quality is still poor. By further combining with our ECB method, the proposed A+ECB method can achieve obviously improvement in both the inter-frame and the intra-frame video qualities.

Furthermore, Table 2 shows the results of a subjective user test experiment. In this experiment, 30 users are asked to view the enhanced videos with different methods and give evaluation scores with a range of 1-5 [8], with 1 for very poor quality and 5 for very good quality. The orders of the enhanced videos are randomly placed and unknown to the users. The scores are averaged over different users and over different video sequences, as shown in Table 2. The subjective evaluation results in Table 2 further demonstrate the effectiveness of our proposed method.

Table 1 Objective measurements comparison for different methods

| | Orig | HEM [4] | LB [8] | ECB | ACB | A+ECB |
|---|---|---|---|---|---|---|
| H | 4.14 | 4.70 | 4.58 | 4.65 | 4.85 | 4.84 |
| HIBTE | 0.292 | 0.285 | 0.331 | 0.248 | 0.309 | 0.249 |
| TAMBE (μ) | 7.69 | 6.36 | 8.84 | 3.35 | 8.54 | 3.41 |
| TAMBE (σ) | 19.16 | 15.63 | 7.74 | 7.24 | 7.51 | 7.19 |

Table 2 User test results for different methods (Note: the scores are averaged over all users and all video sequences)

| | Orig | HEM [4] | LB [8] | ECB | ACB | A+ECB |
|---|---|---|---|---|---|---|
| Average Score | 2.08 | 2.76 | 3.02 | 2.88 | 3.52 | 3.82 |

## IV. Conclusion

In this paper, we propose a new intra-and-inter-constraint-based algorithm for video enhancement. The proposed method analyzes features from different ROIs and creates a 'global' tone mapping curve for the entire frame such that the intra-frame quality of a frame can be properly enhanced. Furthermore, new inter-frame constraints are introduced in the proposed algorithm to further improve the inter-frame qualities among frames. Experimental results demonstrate the effectiveness of our algorithm.

## References

[1] M. Sun, Z. Liu, J. Qiu, Z. Zhang, and M. Sinclair, "*Active lighting for video conferencing,*" *IEEE Trans. Circuits and Systems for Video Technology*, vol.19, no.12, pp.1819-1829, 2009.

[2] G. R. Shorack, "Probability for Statisticians," Springer, 2000.

[3] J. A.Stark, "Adaptive Image Contrast Enhancement using Generalizations of Histogram Equalization," *IEEE Trans. Image Processing*, vol. 9, no. 5, pp. 889 – 896, 2000.

[4] T. Arici, S. Dikbas, Y. Altunbasak, "A Histogram Modification Framework and Its Application for Image Contrast Enhancement," *IEEE Trans. Image Processing*, vol. 18, no. 9, pp. 1921-1935, 2009.

[5] E. Reinhard, M. Ashikhmin, B. Gooch, and P. Shirley, "Color transfer between images," *IEEE Computer Graphics & Appli*, pp. 34-41, 2001.

[6] P. Viola and M. Jones, "Robust real-time object detection," *Second Int'l Workshop on Statistical and Computational Theories of Vision*, 2001.

[7] C. Shi, K. Yu, J. Li, and S. Li, "Automatic image quality improvement for videoconferencing," *ICASSP*, pp. 701-704, 2004.

[8] Z. Liu, C. Zhang and Z. Zhang, "Learning-Based Perceptual Image Quality Improvement for Video Conferencing," *ICME*, pp. 1035-1038, 2007.

[9] W.-C. Chiou and C.-T. Hsu, "Region-Based Color Transfer from Multi-Reference with Graph-Theoretic Region Correspondence Estimation," *ICIP*, pp. 501-504, 2009.

[10] Y.W. Tai, J. Jia and C.K Tang, "Local Color Transfer via Probabilistic Segmentation by Expectation-Maximization," *CVPR*, pp. 747-754, 2005.

[11] Raimondo Schettini and Francesca Gasparini, "Contrast Image Correction Method," *Journal of Electronic Imaging*, vol. 19, no.2, 2010.

[12] S. Battiato and A. Bosco, "Automatic Image Enhancement by Content Dependent Exposure Correction," *EURASIP Journal on Applied Signal Processing*, 2004.

[13] G. D. Toderici and J. Yagnik, "Automatic, Efficient, Temporally-Coherent Video Enhancement for Large Scale Applications," *ACM Multimedia*, pp. 609-612, 2009.

[14] M. Swain and D. Ballard, "Color indexing", *Int'l Journal of Computer Vision*, vol. 7, no. 1, pp. 11-32, 1991.